\pdfoutput=1

\documentclass[11pt]{article}

\usepackage{EMNLP2022}

\usepackage{times}
\usepackage{latexsym}

\usepackage[T1]{fontenc}

\usepackage[utf8]{inputenc}

\usepackage{microtype}

\usepackage{inconsolata}

\usepackage{amsmath,amssymb,amsfonts}
\usepackage{graphicx}
\usepackage{xcolor}
\usepackage{booktabs}
\usepackage{subcaption}
\usepackage{bm}
\usepackage{array}

\title{AFEC: A Knowledge Graph Capturing Social Intelligence in Casual Conversations}

\author{Yubo Xie, Junze Li, and Pearl Pu \\
  School of Computer and Communication Sciences \\
  \'{E}cole Polytechnique F\'{e}d\'{e}rale de Lausanne\\
  Lausanne, Switzerland \\
  \texttt{\{yubo.xie,\,junze.li,\,pearl.pu\}@epfl.ch} \\}

\begin{document}
\maketitle
\begin{abstract}
This paper introduces AFEC, an automatically curated knowledge graph based on people's day-to-day casual conversations. The knowledge captured in this graph bears potential for conversational systems to understand how people offer acknowledgement, consoling, and a wide range of empathetic responses in social conversations. For this body of knowledge to be comprehensive and meaningful, we curated a large-scale corpus from the \texttt{r/CasualConversation} SubReddit.
After taking the first two turns of all conversations, we obtained 134K speaker nodes and 666K listener nodes. To demonstrate how a chatbot can converse in social settings, we built a retrieval-based chatbot and compared it with existing empathetic dialog models. Experiments show that our model is capable of generating much more diverse responses (at least 15\% higher diversity scores in human evaluation), while still outperforming two out of the four baselines in terms of response quality.
\end{abstract}

\section{Introduction}
\begin{figure}[t]
    \centering
    \includegraphics[width=\columnwidth]{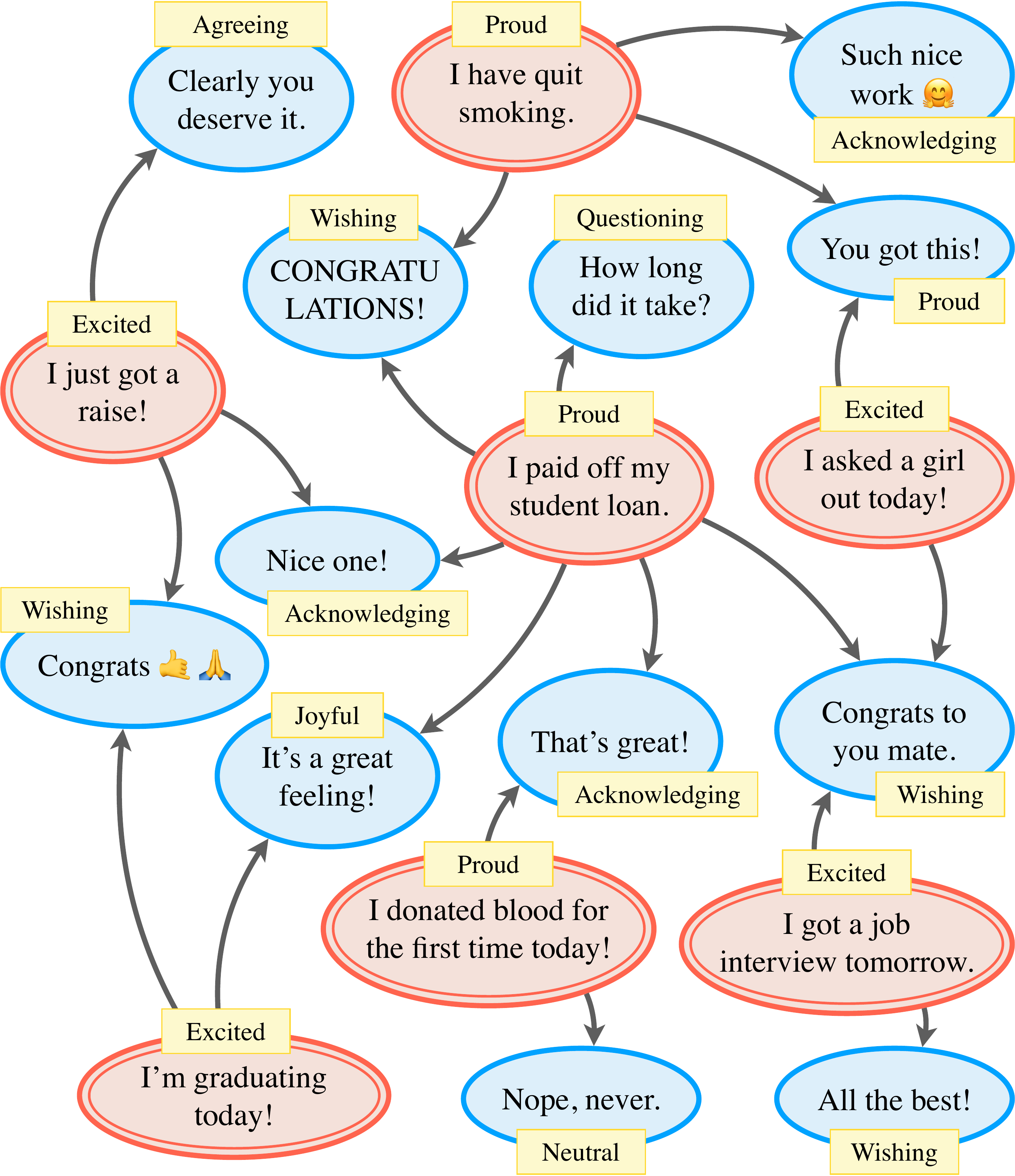}
    \caption{A snippet of AFEC, our knowledge graph of social intelligence in casual conversations. Red nodes represent speaker utterances, which start a conversation, and blue nodes represent the corresponding listener utterances, labeled with the desired emotions necessary for continuing the conversation.}
    \label{fig:snippet}
\end{figure}
Social intelligence is our ability to optimally understand our social environment and react wisely~\cite{ganaie2015study}. As one type of commonsense knowledge, social intelligence encompasses various abilities such as emotion reading, empathy, and knowledge of social rules. According to \citet{daniel2006social}, social intelligence consists of two ingredients: \emph{social awareness}, the ability of understanding others' feelings and grasping the complexity of social situations, and \emph{social facility}, the ability of making smooth and effective interactions based on social awareness. For example, if someone says ``my job interview didn't go well'', one could immediately sense that the person is upset and might be seeking consolation. Based on such social awareness, a good way to respond could be ``don't give up and I'm sure you'll do better next time''. Such examples of applying social intelligence exist everywhere in our daily life and can easily be found in human-to-human conversations.

Having desirable social intelligence does not always seem easy and straightforward even for human beings, not to mention AI systems.
Previous work has attempted at creating knowledge graphs to explicitly represent different commonsense knowledge so that AI systems could utilize them as external resources to perform various types of commonsense reasoning. This ranges from knowledge about common concepts~\cite{DBLP:conf/aaai/SpeerCH17,DBLP:conf/acl/TandonMW17} to inferences over common events~\cite{DBLP:conf/aaai/SapBABLRRSC19,DBLP:conf/www/ZhangLPSL20}. However, to the best of our knowledge, there is still no work on building a knowledge graph which captures the social intelligence that people display when conducting social conversations with each other in day-to-day social environments. By inspecting the knowledge graph, one could learn how to appropriately respond to another person with desired emotions and intents. The knowledge graph can also serve as an external resource for the development of both open-domain and task-oriented dialog systems.

In this paper, we present \textbf{AFEC} (\textbf{AF}fable and \textbf{E}ffective \textbf{C}onverser), an automatically curated knowledge graph that captures social intelligence in casual conversations between real people on various topics commonly found in daily life. Figure~\ref{fig:snippet} gives a snippet of our knowledge graph, showing different ways people respond to some experience shared by another person. Due to the limitations of existing open-domain dialog datasets (large-scale datasets are usually noisy, and manually labeled datasets are small in size), we decided to manually crawl conversational data from the \texttt{r/CasualConversation} Subreddit,\footnote{\url{https://www.reddit.com/r/CasualConversation/}} which covers a wide range of topics in people's day-to-day social communication. After preprocessing, we grouped and merged the utterances into nodes of the knowledge graph. Each node is a short utterance, with the red ones representing the speaker (usually sharing some past experience) and the blue ones representing the listener (responding to the speaker). Following the taxonomy of empathetic response intents proposed by \citet{DBLP:conf/coling/WelivitaP20}, we labeled each node with one of the 41 categories of emotions/intents, by training a Transformer~\cite{DBLP:conf/nips/VaswaniSPUJGKP17} classifier on manually labeled dialog data. Unlike other commonsense knowledge graphs that deal with concepts or events, ours focuses on verbal skills to demonstrate empathy, which are ubiquitous in people's daily social environments, and was curated automatically from data, without the efforts of manual selecting and labeling.

Our contributions are as follows: (1) We designed a completely automatic curation pipeline to create a knowledge graph of social intelligence covering people's daily communication, which can be used as an external resource to improve the performance of dialog systems; (2) As a by-product of our knowledge graph, we obtained a large-scale casual conversation dataset that has a good trade-off between size and quality; the knowledge graph and the dataset will be publicly available soon; (3) We designed a simple retrieval-based chatbot using the knowledge graph, without the efforts of training a neural network, and the comparison with existing empathetic dialog models showed that our retrieval model is capable of generating much more diverse responses, yet still producing quality responses.

\begin{figure*}[t]
    \centering
    \includegraphics[width=0.9\textwidth]{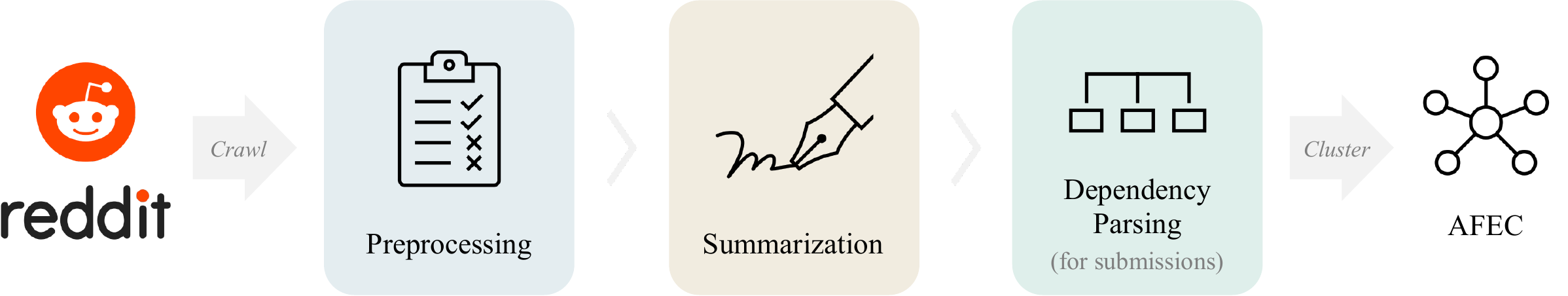}
    \caption{The overall workflow of the data curation process.}
    \label{fig:workflow}
\end{figure*}

\section{Related Work}
\paragraph{Commonsense Knowledge}
It is believed that \citet{DBLP:journals/ac/Bar-Hillel60} was the first to mention the importance of incorporating commonsense knowledge into natural language processing systems, in the context of machine translation. Broadly speaking, there are two types of commonsense knowledge. One type of commonsense knowledge that humans generally acquire is ``naive physics'', which involves inference of how physical objects interact with each other. For example, if one is told that a glass of water falls onto the floor, he/she will most likely infer that the glass shatters and the floor becomes wet. Another type of commonsense knowledge that humans have is ``intuitive psychology''. This type of knowledge enables us to infer people's behaviors, intents, or emotions. For example, one could easily tell that a person who has lost his/her job probably feels upset. There is abundant recent work that explicitly represents commonsense knowledge into a structured knowledge graph.
ConceptNet~\citep{DBLP:conf/aaai/SpeerCH17} is a directed graph whose nodes are concepts, and the edges represent assertions of commonsense about the concepts, e.g., \emph{IsA}, \emph{IsUsedFor}, \emph{MotivatedByGoal}, etc. The nodes are natural language phrases, e.g., noun phrases, verb phrases, or clauses. The latest version is ConceptNet 5.5, which contains over 8 million nodes and over 21 million links. SenticNet~\citep{DBLP:conf/cikm/CambriaLXPK20} incorporates a set of semantics, sentics, and polarity associated with 200,000 natural language concepts. Specifically, semantics define the denotative information associated with natural language phrases, sentics define the emotion categorization values (expressed in terms of four affective dimensions) associated with these concepts, and polarity is floating number between $-1$ and $+1$. WebChild~\citep{DBLP:conf/acl/TandonMW17} is a large-scale commonsense knowledge graph that was automatically extracted and disambiguated from Web contents, using semi-supervised label propagation over graphs of noisy candidate assertions. It contains triples that connect nouns with adjectives via fine-grained relations such as \emph{hasShape}, \emph{hasTaste}, \emph{evokesEmotion}, etc. The newest version WebChild 2.0 was released in 2017 and contains over 2 million concepts with 18 million assertions. \textsc{Atomic}~\citep{DBLP:conf/aaai/SapBABLRRSC19} is a commonsense knowledge graph consisting of 877K textual descriptions of inferential knowledge obtained from crowdsourcing. It focuses on \emph{if-then} relations between events and possible inferences over the events. The base events were extracted from a variety of corpora including stories and books. The \textsc{Atomic} knowledge graph contains a total number of 309,515 nodes and 877,108 triples. ASER~\citep{DBLP:conf/www/ZhangLPSL20} is a large-scale eventuality knowledge graph automatically extracted from more than 11-billion-token unstructured textual data. It contains 15 relation types belonging to five categories, 194 million unique eventualities, and 64 million edges between them. The eventualities were extracted from a wide range of corpora from different sources, according to a selected set eventuality patterns. The eventuality relations were also automatically extracted using a selected set of seed connectives.

\paragraph{Emotional Dialog Data}
There are not many emotional dialog datasets that are publicly available, and most of them are limited in size. \citet{DBLP:conf/ijcnlp/LiSSLCN17} created the DailyDialog dataset from English learning websites, consisting of 13K multi-turn dialogs manually labeled with 7 emotions. The EmotionLines dataset~\cite{DBLP:conf/lrec/HsuCKHK18} contains 2,000 dialogs collected from Friends TV scripts and EmotionPush chat logs, labeled with 7 emotions. \citet{DBLP:conf/semeval/ChatterjeeNJA19} proposed the EmoContext dataset collected from users' interaction with a conversational agent, which contains 38K dialogs labeled with 4 emotions. \citet{DBLP:conf/acl/RashkinSLB19} curated the EmpatheticDialogues dataset containing 25K dialogs collected from a crowdsourcing platform by letting workers communicate with each other based on 32 emotion categories. \citet{DBLP:conf/emnlp/WelivitaXP21} adopted a semi-supervised approach to label the OpenSubtitles dialog data, and obtained a large-scale dialog dataset EDOS that contains 1M empathetic dialogs. However, the dialogs curated from movie subtitles are often noisy and don't fully overlap with daily social topics. Therefore, it is desirable to find conversations from sources that fit more with daily social environments.

\section{Data Curation}
Since our goal is to build a knowledge graph of social intelligence from people's everyday social interactions, it is necessary to have a large-scale daily dialog dataset. Existing dialog datasets (that are publicly available) mainly falls into two categories: task-oriented dialogs and open-domain dialogs. Task-oriented dialogs are extracted from some specific domain, for example the Ubuntu Dialogue Corpus~\cite{DBLP:conf/sigdial/LowePSP15}, which includes technology related conversations mostly. These dialogs do not fit into the day-to-day social settings, thus not applicable to our case. Open-domain dialogs, on the other hand, covers a much broader range of topics in daily life, thus more suitable for our purpose. However, existing open-domain dialog datasets have either low quality or limited size. For example, dialog datasets created from movie subtitles, for example OpenSubtitles~\cite{DBLP:conf/lrec/LisonTK18}, often have a large scale and a diverse range of topics, but the separation between two conversation scenes is hard to detect precisely, and the text quality cannot always be guaranteed due to transcription errors. On the other hand, open-domain dialog datasets of high quality, for example EmpatheticDialogues~\cite{DBLP:conf/acl/RashkinSLB19}, are often limited in size, because they are created manually or through crowdsourcing.

Due to the limitations of existing dialog datasets, we would like to curate a large-scale dialog dataset containing high quality casual conversations in daily life, by crawling from online resources. The overall workflow is illustrated in Figure~\ref{fig:workflow}. Next we are going to describe each step in detail.

\paragraph{Crawling from Reddit}
Reddit\footnote{\url{https://www.reddit.com/}} is a social news, rating and discussion website, where users can post what they want to share in subreddits with different themes. Among all subreddits, \texttt{r/CasualConversation} is a subreddit for users to talk about common topics, like ``\emph{Today I finish to pay to my loan}'' and ``\emph{I'm starting a new job today}''. We used the Pushshift Reddit API\footnote{\url{https://pushshift.io/}} to crawl submission and comment posts on \texttt{r/CasualConversation}. Limiting the publishing time from January 1, 2016 to December 31, 2021, we obtained 387,594 submissions and 4,152,652 comments in total. Among all the comments, there are 1,908,867 comments that directly reply to submission posts. To form the conversations, the submissions are considered as the speaker utterances and we only selected the comments that directly reply to submissions as the listener utterances.

\paragraph{Preprocessing}
To preprocess all the crawled text, we design the following rules: 
\begin{enumerate}
    \item We replace the HTML escape characters with their normal ones (e.g., \texttt{\&gt;} becomes \texttt{>});
    \item We remove any content in brackets;
    \item We remove redundant spaces (including breaklines) in the input text;
    \item We discard the input text if it is empty;
    \item We discard the input text if it is ``[deleted]'' or ``[removed]'';
    \item We discard the input text if it contains URL;
    \item We discard the input text if it contains ``r/<subreddit>'', ``u/<username>'' or ``reddit'';
    \item We discard the input text if the percentage of alphabetical letters is less than 70\%;
    \item We discard the input text if the number of tokens is less than 2.
\end{enumerate}

\paragraph{Summarization}
After preprocessing, there still exist some long utterances (for example some submissions have one or more blocks of text) containing multiple sentences. To keep the crawled conversations succinct, we applied the SMMRY algorithm\footnote{\url{https://smmry.com/about}} to summarize these utterances into one sentence. 

\paragraph{Dependency Parsing}
We observed that a submission usually consists of two parts: a title and possibly a block of description text further explaining the situation, and in most cases the title itself is enough to summarize what the speaker is going to express. However, we need to deal with the cases where the title is just a phrase and does not fully describe the whole submission. Therefore, we would like to check if a speaker utterance indicates some specific actions, propositions or statements. Specifically, we used the SpaCy\footnote{\url{https://spacy.io/}} package to generate a dependency parsing tree, and if the root of the tree is a verb, we kept the sentence as the speaker utterance. Note that we first applied this procedure on the title (i.e., first try summarizing the title and then apply dependency parsing), and if the final result was discarded, we then applied the same procedure on the description text.

\paragraph{Clustering}
After looking into the the utterances filtered by previous steps, we observed that some speaker utterances are similar in meaning (the same case for listener utterances). Therefore, these similar utterances can be clustered into one node. We first encoded each utterance into a vector with dimension 768 using Sentence-BERT~\cite{DBLP:conf/emnlp/ReimersG19}, and then calculated the cosine similarity score between vector representations of two utterances. If the similarity value between two speaker utterances is more than 0.85 (and 0.80 for listener utterances), they are grouped together. We used a fast community detection algorithm\footnote{\url{https://www.sbert.net/examples/applications/clustering/README.html\#fast-clustering}} to generate all the speaker nodes and listener nodes. However, due to the number of listener utterances being too large, the algorithm does not fit into the memory (256GB). Thus, we evenly split the listener utterances into two parts and applied the algorithm to find the respective clusters, and then ran another round of community detection on the obtained clusters. To connect these nodes into one graph, an edge is added between a speaker node and a listener node, if at least one of the utterances in the speaker node matches one of the utterances in the listener node. In this step, we only considered utterances with length not greater than 40.

\vspace{0.5\baselineskip}
Finally, we combined all the selected dialogs crawled from Reddit and the dialogs from EmpatheticDialogues dataset. There are 152,680 speaker utterances and 838,785 listener utterances in total. After clustering, we have \textbf{134,061 speaker nodes} and \textbf{666,587 listener nodes}.

\section{Node Labeling}
When conducting social conversations, people have different patterns of responding to the other interlocutor, which are mostly driven by their emotional states and emotional intelligence. For example, empathy is an important aspect of social conversations, and depending on the other interlocutor's utterance, different people would have different ways to respond to certain emotional events (either joyful or sad), e.g., encouraging, consoling, or questioning. To this end, it is necessary to study the emotions and intents embedded in the utterances in our knowledge graph. To do this, we adopted the taxonomy of empathetic response intents proposed by~\citet{DBLP:conf/coling/WelivitaP20}, which extends the 32 emotion categories of EmpatheticDialogues~\cite{DBLP:conf/acl/RashkinSLB19} with 8 fine-grained empathetic response intents, plus the \emph{neutral} category. We then labeled each node in the graph with one of the 41 emotions/intents (listed in Table~\ref{tab:taxonomy}).
\begin{table}[t]
    \centering
    \begin{tabular}{m{1.5cm}m{5.3cm}}
        \toprule
        \textbf{Emotion} & prepared, anticipating, hopeful, proud, excited, joyful, content, caring, grateful, trusting, confident, faithful, impressed, surprised, terrified, afraid, apprehensive, anxious, embarrassed, ashamed, devastated, sad, disappointed, lonely, sentimental, nostalgic, guilty, disgusted, furious, angry, annoyed, jealous \\
        \midrule
        \textbf{Intent} & agreeing, acknowledging, encouraging, consoling, sympathizing, suggesting, questioning, wishing, neutral\\
        \bottomrule
    \end{tabular}
    \caption{Taxonomy of emotions and intents.}
    \label{tab:taxonomy}
\end{table}

\begin{figure}[t]
    \centering
    \includegraphics[width=\columnwidth]{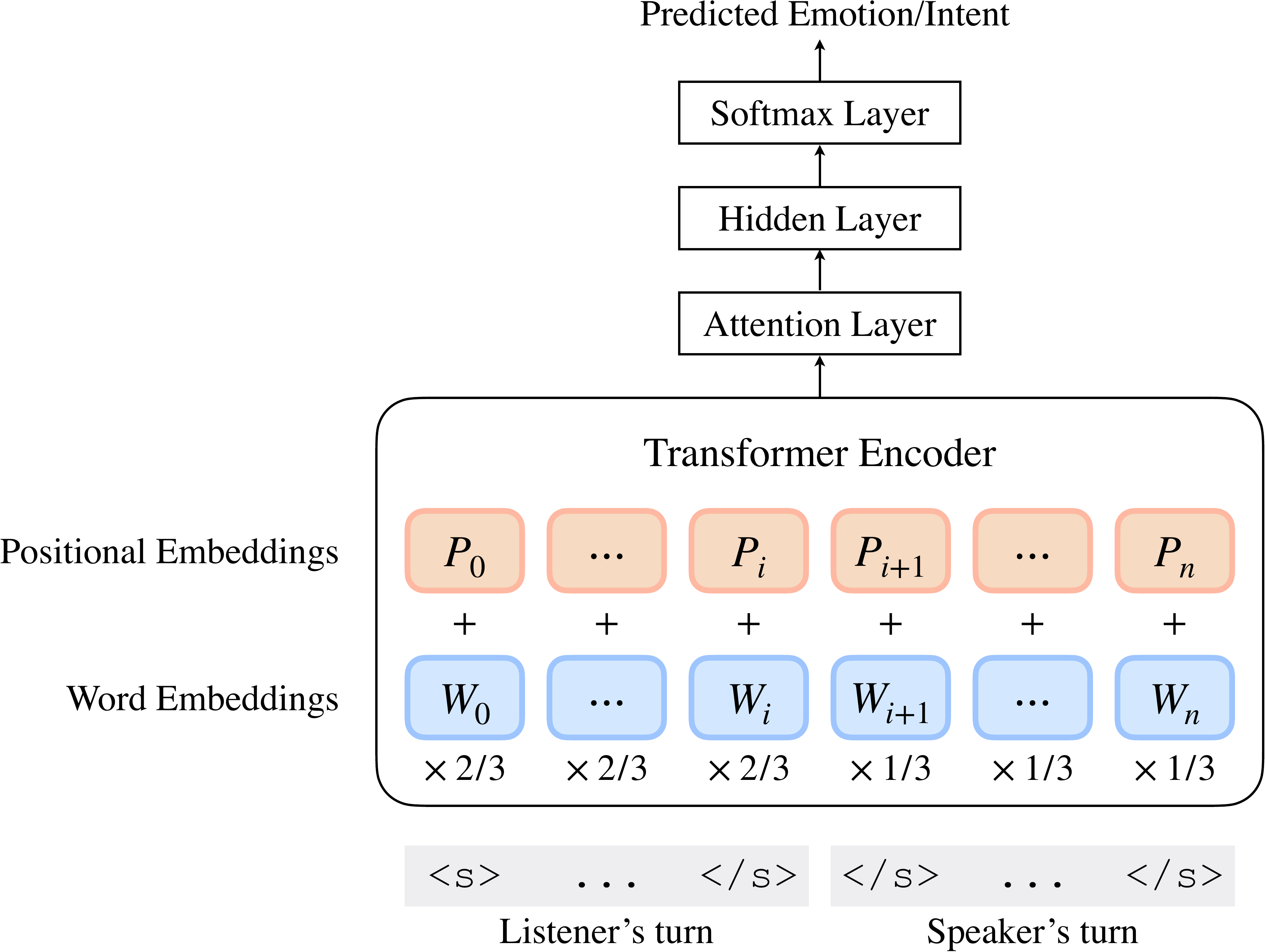}
    \caption{Architecture of the emotion/intent classifier.}
    \label{fig:classifier_architecture}
\end{figure}

Specifically, we followed the work of \citet{DBLP:conf/emnlp/WelivitaXP21} and trained the same emotion/intent classifier on the dialog data labeled with crowdsourcing workers and then extended with distant learning (14K in total). Figure~\ref{fig:classifier_architecture} depicts the architecture of the classifier. We adopted a Transformer encoder architecture with 12 layers, 768 hidden units, 12 multi-heads, and initialized the model with weights from the pre-trained language model RoBERTa~\cite{DBLP:journals/corr/abs-1907-11692}. When labeling the speaker's turn (red nodes in the knowledge graph), we just feed the speaker's utterance as input into the model. When labeling the listener's turn (blue nodes in the knowledge graph), we append the corresponding speaker's utterance after the listener's utterance and apply a decaying weight factor so that the model pays more attention to the listener's utterance (as shown in Figure~\ref{fig:classifier_architecture}).

\begin{figure*}[t]
    \centering
    \begin{subfigure}[t]{\textwidth}
        \centering
        \includegraphics[width=\textwidth]{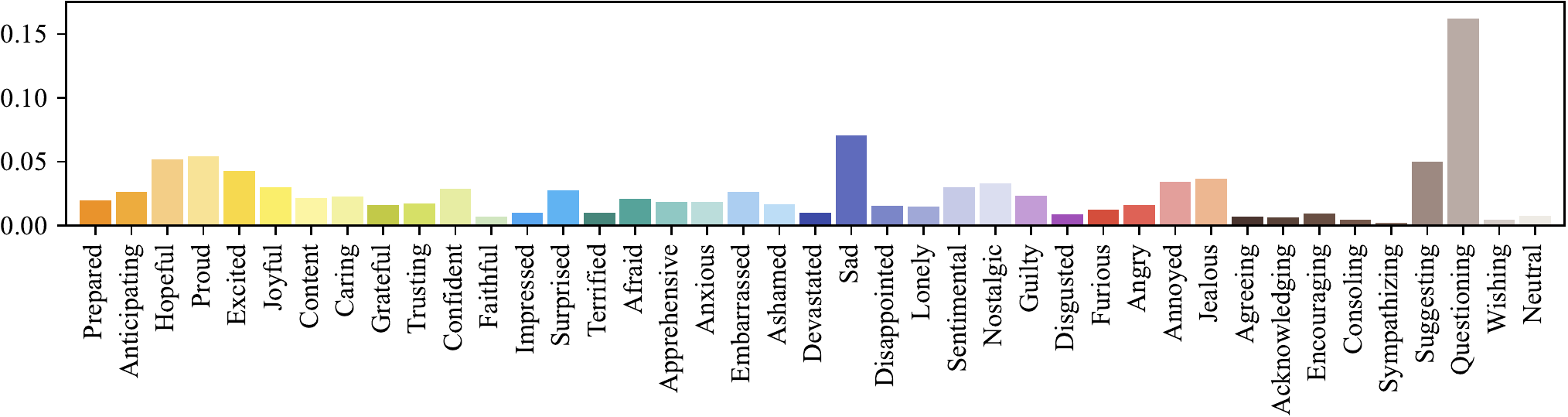}
        \caption{Distribution of emotion/intent in the speaker nodes.}
        \label{fig:emot_dist_speaker}
    \end{subfigure}\\[1em]
    \begin{subfigure}[t]{\textwidth}
        \centering
        \includegraphics[width=\textwidth]{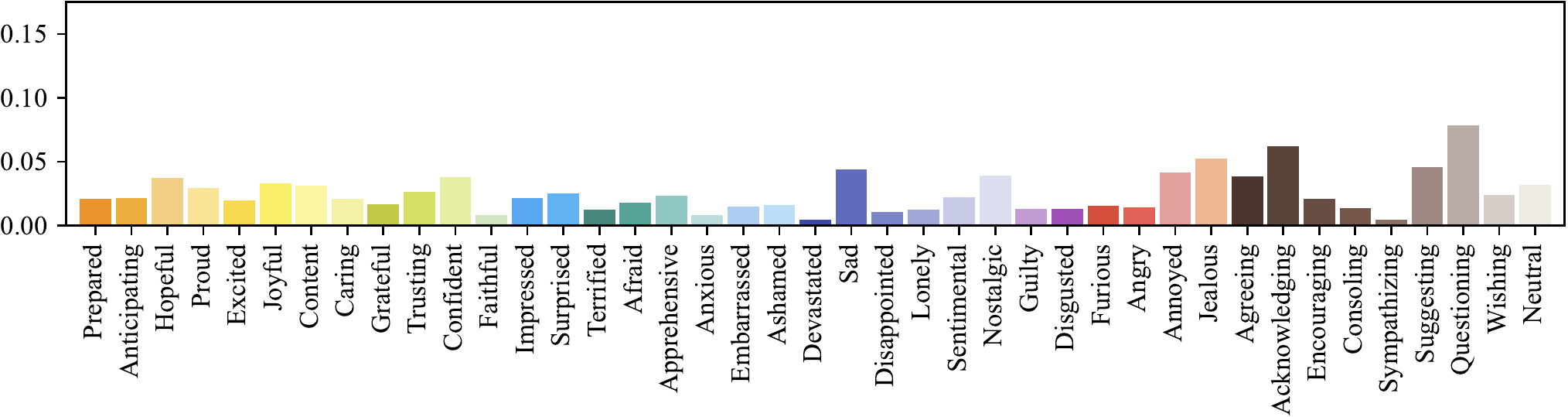}
        \caption{Distribution of emotion/intent in the listener nodes.}
        \label{fig:emot_dist_listener}
    \end{subfigure}
    \caption{Distributions of emotion/intent in AFEC.}
    \label{fig:emot_dist}
\end{figure*}

After obtaining the labels for each node, we visualized the distribution of the emotion/intent of the speaker nodes (Figure~\ref{fig:emot_dist_speaker}) and that of the listener nodes (Figure~\ref{fig:emot_dist_listener}), respectively. We make the bars of the emotion categories more colorful than the intent categories. It can be observed that the listener nodes are generally less emotional than the speaker nodes, and for the listener nodes, the empathetic response intents, such as \emph{agreeing}, \emph{acknowledging}, \emph{encouraging}, \emph{consoling}, \emph{sympathizing}, and \emph{wishing}, are much more prominent than the speaker nodes. This is also consistent with the findings of \citet{DBLP:conf/coling/WelivitaP20}.

\section{Experiments}

Our curated casual conversational graph can serve as a good data source for the development of open-domain dialog models. In this section, we show how we can build a simple retrieval-based dialog model using the conversation graph, which is capable of generating very diverse and human-like responses, yet still achieves better performance than some of the existing generative dialog models, using both automatic and human evaluation.

\subsection{Data Split}
In order to evaluate the retrieval-based chatbot as well as the baselines, we split out roughly 10\% of all the utterances in the knowledge graph, reserved as the testing set for our following experiments. Specifically, we first selected out all the speaker utterances that come from the EmpatheticDialogues testing set, and then, in the remaining speaker utterances, we randomly chose 10\% and combined them with the previously selected EmpatheticDialogues utterances, to form the final testing set. As a result, the testing set consists of 15,212 speaker utterances and 82,284 listener utterances.

\subsection{A Retrieval-Based Dialog Model}
Based on the casual conversation knowledge graph, we are able to develop a simple retrieval-based dialog model, AFEC-Talk, by comparing the input utterance with the speaker nodes in the graph and then return the replies (listener nodes) that correspond to the most similar speaker node. In particular, for an input utterance $x$, we first encode it into a vector with dimension 768 using Sentence-BERT, which is denoted by $\bm{v}_x$. We then calculate the cosine similarity score between $\bm{v}_x$ and $\bm{v}_s$, where $\bm{v}_s$ is the vector representation (also encoded using Sentence-BERT) of a speaker node $s$ in the graph. By sorting all the speaker nodes by their cosine similarity scores with $x$, we pick the most similar one, denoted by $\hat{s}$. Given $\hat{s}$, we can find all its corresponding listener nodes $L_{\hat{s}}$, among which we are going to select one utterance as the dialog model's response. We have the following strategies for selecting the reply:
\begin{enumerate}
\item \textbf{Select randomly}. We just randomly select a listener utterance from $L_{\hat{s}}$.
\item \textbf{Select the reply with the highest degree}. We select the listener utterance that has the most in-going edges, meaning that this utterance is more frequently used as a reply than other utterances in the graph.
\item \textbf{Select the reply that follows the emotion of the input utterance}. We simply select the reply that has the same or similar emotion/intent of the input utterance. If such reply can not be found, we just randomly pick one from $L_{\hat{s}}$. We consider emotions/intents to be similar if they belong to the same group, as defined in Appendix~\ref{sec:sim_emos_intents}.
\item \textbf{Select the reply with empathetic response intents}. We select the reply that is labeled with one of the 8 empathetic response intents (excluding \emph{neutral}).
\end{enumerate}
For all the strategies, if multiple candidate replies exist, we would randomly select one.

\subsection{Baselines}
Empathy is an important aspect of people's daily social communication, and is considered as an essential ability of open-domain chatbots. To this end, we compare our retrieval-based dialog model with some existing empathetic open-domain dialog models. Due to limited retrieval-based chatbots using knowledge graphs to generate empathetic responses, we decided to compare AFEC-Talk with existing end-to-end generative systems. Our baselines are:
\begin{itemize}
\item \textbf{MoEL}~\cite{DBLP:conf/emnlp/LinMSXF19}. MoEL is an end-to-end empathetic dialog model that uses multiple decoders (listeners) to react to each context emotion accordingly. According to the emotion classification distribution, a meta-listener then combines the output states of each listener, to generate the final empathetic response.
\item \textbf{MIME}~\cite{DBLP:conf/emnlp/MajumderHPLGGMP20}. MIME generates empathetic responses by exploiting the assumption that an empathetic conversational agent would often mimic the user's emotion to a certain degree, depending on whether it is positive or negative. The model also introduces some stochasticity into the emotion mixture to generate more varied responses.
\item \textbf{CEM}~\cite{DBLP:journals/corr/abs-2109-05739}. When generating the empathetic response, CEM first queries COMET~\cite{DBLP:conf/acl/BosselutRSMCC19}, a GPT-2~\cite{radford2019language} based model fine-tuned on ATOMIC, to get the commonsense inferences of the input context, and then uses a knowledge selector to fuse the obtained information.
\item \textbf{BlenderBot}~\cite{DBLP:conf/eacl/RollerDGJWLXOSB21}. BlenderBot was pre-trained on Reddit data and then fine-tuned on the BlendedSkillTalk dataset~\cite{DBLP:conf/acl/SmithWSWB20}, which combines conversation skills from three individual dialog datasets that focus on engaging personality, empathy, and knowledge, respectively. We use the 90M model that is publicly available.
\item \textbf{MEED2}~\cite{DBLP:conf/conll/XieP21}. MEED2 incorporates extra fine-grained empathetic intents into the emotion distribution, and uses a separate Transformer encoder to determine the emotion/intent of the response to be generated. The model was pre-trained on OpenSubtitles dialogs and fine-tuned on smaller dialog datasets. We use the version of MEED2 that was fine-tuned on the EmpatheticDialogues dataset.
\end{itemize}

\subsection{Automatic Evaluation}

\begin{table*}[t]
\centering
\begin{tabular}{lccccccc}
\toprule
\textbf{Model} & \textbf{BLEU-2} & \textbf{BLEU-4} & \textbf{ROUGE-2} & \textbf{METEOR} & \textbf{Dist-1} & \textbf{Dist-2} & \textbf{Dist-3} \\
\midrule
MoEL       & 0.0232 & 0.0016 & 0.0040 & 0.1098 & 0.0021 & 0.0092 & 0.0191 \\
MIME       & 0.0215 & 0.0011 & 0.0034 & 0.1298 & 0.0011 & 0.0042 & 0.0074 \\
CEM        & 0.0184 & 0.0013 & 0.0032 & 0.0862 & 0.0021 & 0.0080 & 0.0151 \\
BlenderBot & 0.0383 & 0.0022 & 0.0054 & \textbf{0.1615} & 0.0151 & 0.0873 & 0.1778 \\
MEED2      & 0.0300 & 0.0025 & 0.0047 & 0.0972 & 0.0141 & 0.0681 & 0.1314 \\
\midrule
$\text{AFEC-Talk}_\textit{rand}$   & 0.0475 & \textbf{0.0223} & 0.0110 & 0.1294 & 0.0702 & \textbf{0.4527} & \textbf{0.7915} \\
$\text{AFEC-Talk}_\textit{hd}$     & \textbf{0.0528} & 0.0181 & \textbf{0.0124} & 0.1274 & 0.0703 & 0.4286 & 0.7363 \\
$\text{AFEC-Talk}_\textit{follow}$ & 0.0464 & 0.0223 & 0.0110 & 0.1275 & 0.0696 & 0.4433 & 0.7695 \\
$\text{AFEC-Talk}_\textit{intent}$ & 0.0481 & 0.0204 & 0.0121 & 0.1257 & \textbf{0.0703} & 0.4381 & 0.7592 \\
\bottomrule
\end{tabular}
\caption{Automatic evaluation results. Dist-$n$ denotes Distinct-$n$ score. $\text{AFEC-Talk}_{*}$ denotes our retrieval-based dialog model with different reply selecting strategies: \textit{rand} means selecting randomly; \textit{hd} means selecting the reply with the highest degree; \textit{follow} means following the input emotion/intent; \textit{intent} means selecting the reply with one of the 8 empathetic response intents.}
\label{tab:auto_eval}
\end{table*}

\paragraph{Metrics}
To automatically evaluate AFEC-Talk as well as the baselines, we use metrics that are widely adopted for the evaluation of dialog models:
\begin{itemize}
\item \textbf{BLEU}~\cite{DBLP:conf/acl/PapineniRWZ02}. BLEU compares the model output with a golden answer by counting the matching $n$-grams and calculating the precision score. We use cumulative BLEU scores BLEU-2 and BLEU-4.
\item \textbf{ROUGE}~\cite{lin2004rouge}. ROUGE measures the match rate of $n$-grams between the model output and the reference by calculating precision, recall, and F1 scores.
\item \textbf{METEOR}~\cite{DBLP:conf/acl/BanerjeeL05}. METEOR combines unigram precision and recall using a harmonic mean, with recall weighted more than precision. Additionally, it computes a penalty using longer $n$-gram matches.
\item \textbf{Distinct}~\cite{DBLP:conf/naacl/LiGBGD16}. Distinct-$n$ measures the diversity of the generated responses by calculating the ratio of unique $n$-grams over the total number of $n$-grams in the generated responses.
\end{itemize}
Note that in the testing set, a speaker utterance could have multiple corresponding listener utterances. In this case, for metrics that were not designed for multiple references, we simply average the individual scores.

\paragraph{Results}
Table~\ref{tab:auto_eval} shows the results of automatic evaluation. As shown in the table, our retrieval-based dialog model AFEC-Talk (with different reply selecting strategies) outperforms the baselines on most of the metrics. Notably, our model has higher BLEU and ROUGE scores than the baselines, and has significantly higher Distinct scores than the baselines. This is because generative dialog models notoriously suffer from the problem of generated responses being generic and repetitive, while AFEC-Talk directly pulls candidate replies from the casual conversation knowledge graph, which are much more diverse and interesting. In particular, it is also intuitive that the strategy of randomly selecting a reply achieves the highest Distinct scores. We also notice that, among all the baselines, BlenderBot and MEED2 generally perform better. We reckon this is because they were pre-trained on large-scale datasets such as Reddit and OpenSubtitles dialogs, while other baselines were just trained on the EmpatheticDialogues dataset, which is much smaller.

\subsection{Human Evaluation}

\begin{table*}[t]
\centering
\begin{tabular}{lccccccc}
\toprule
\textbf{Model} & \textbf{Good (\%)} & \textbf{Okay (\%)} & \textbf{Bad (\%)} & \textbf{Avg.~Score} & \textbf{Diversity} \\
\midrule
MIME       & 11.50 & 16.50 & 72.00 & 0.3950 & 2.1250 \\
CEM        & 19.50 & 24.00 & 56.50 & 0.6300 & 2.5000 \\
BlenderBot & 40.00 & 16.50 & 43.50 & 0.9650 & 3.0750 \\
MEED2      & \textbf{44.00} & 28.00 & 28.00 & \textbf{1.1600} & 2.8500 \\
\midrule
$\text{AFEC-Talk}_\textit{rand}$ & 32.00 & 21.50 & 46.50 & 0.8550 & \textbf{3.5500} \\
$\text{AFEC-Talk}_\textit{hd}$   & 27.00 & 25.50 & 47.50 & 0.7950 & 3.4000 \\
\bottomrule
\end{tabular}
\caption{Human evaluation results. $\text{AFEC-Talk}_{*}$ denotes our retrieval-based dialog model with different reply selecting strategies: \textit{rand} means selecting randomly; \textit{hd} means selecting the reply with the highest degree.}
\label{tab:human_eval}
\end{table*}

\paragraph{Set-up}
For dialog generation, an input utterance could have multiple responses that are equally good, so mere automatic evaluation is not enough. To this end, we designed a human experiment to evaluate the models. We randomly selected 200 dialogs (each including one speaker utterance and possibly multiple listener utterances) from the testing set, and then recruited 8 workers from Upwork\footnote{\url{https://www.upwork.com/}} to evaluate them.\footnote{We also launched the same experiment on Amazon Mechanical Turk (MTurk), but were not able to recruit enough qualified workers to evaluate all the dialogs. We also compared the results from Upwork and MTurk, and found that the Upwork workers were more attentive to the tasks, and provided more quality answers.} The 200 dialogs are split into 40 batches, with each batch containing 5 dialogs. We then asked each worker to evaluate 5 batches. For each batch, we ask the workers to rate six models (MIME, CEM, BlenderBot, MEED2, $\text{AFEC-Talk}_\textit{rand}$, and $\text{AFEC-Talk}_\textit{hd}$) by dragging and dropping them into three areas, namely \emph{good}, \emph{okay}, and \emph{bad}, according to whether their responses are semantically coherent and emotionally appropriate following the speaker's utterance. In addition to the evaluation of individual responses, we also asked the workers to rate the diversity of each model with a 5-point Likert scale, by showing them all the 5 dialogs at the end of each batch. We paid \$5 to a worker for completing one batch.

\paragraph{Results}
The results of human evaluation are given in Table~\ref{tab:human_eval}. We calculated the percentage of each model being rated with \emph{good}, \emph{okay}, and \emph{bad}. Regarding \emph{good} as 2, \emph{okay} as 1, and \emph{bad} as 0, we also calculated an average score for each model. From the table, we can see that our retrieval-based dialog model AFEC-Talk achieves the highest diversity score, which is consistent with the results of automatic evaluation. In particular, the strategy of randomly selecting a reply performs better than selecting the reply with highest degree. In terms of response quality, AFEC-Talk outperforms both MIME and CEM. However, it does not outperform BlenderBot and MEED2, which we think is because these two models were pre-trained on large-scale dialog datasets. MEED2 achieved the highest average score, and while inspecting the responses generated by MEED2, we found that it tends to generate questions, which is favored by the crowdsourcing workers. This indicates that questioning could be a good strategy of replying to a speaker's utterance, because it enables the chatbot to sound more attentive and show interest in what the speaker has said~\cite{DBLP:conf/coling/WelivitaP20}, and asking a question could guide the speaker to elaborate on the topic, thus further expanding the conversation. Nevertheless, our retrieval-based model still achieves close average scores to BlenderBot and MEED2, despite the fact that our model completely spares any training process.

\section{Conclusion}
In this paper, we present a knowledge graph, AFEC, that captures social intelligence in day-to-day casual conversations. We crawled submissions and their comments from the \texttt{r/CasualConversation} Subreddit, which cover a wide range of daily social topics. After preprocessing and cleaning, the speaker and listener utterances were clustered into the nodes in the knowledge graph. To better understand the emotion and intent of the speaker and listener, we trained a classifier and labeled each node in the knowledge graph with one of the 41 emotions/intents. The resultant knowledge graph contains a total number of 134K speaker nodes and 666K listener nodes. We designed a retrieval-based chatbot, AFEC-Talk. Both offline and human evaluations show that AFEC-Talk can generate highly intelligent social chitchat. Compared with its counterparts that use end-to-end methods, it is more diverse, overcoming one of the long-standing issues in neural generative approaches. As future work, we plan to utilize the knowledge graph for the training of a generative dialog model. We also plan to extend the dialogs in the knowledge graph to multiple turns.

\section{Acknowledgement}
This research was supported by the Swiss National Science Foundation (Grant No.~200021\_184602). We thank Yufan Ren, our colleague at EPFL, for taking time to evaluate the pilot dialogs and validate the online human evaluation process.

\bibliography{references}
\bibliographystyle{acl_natbib}

\appendix
\section{Groups of Similar Emotions/Intents}
\label{sec:sim_emos_intents}
For selecting the reply that follows the emotion or intent of the input utterance, we defined 20 groups of similar emotions and intents, according to our intuition. See Table~\ref{tab:group_sim_emo_intent}.
\begin{table}[ht]
    \centering
    \begin{tabular}{cp{6cm}}
        \toprule
        \textbf{\#} & \textbf{Emotion/Intent} \\
        \midrule
        1 & prepared, confident, proud \\
        2 & content, hopeful, anticipating \\
        3 & joyful, excited \\
        4 & caring \\
        5 & faithful, trusting, grateful \\
        6 & jealous, annoyed, angry, furious \\
        7 & terrified, afraid, anxious, apprehensive \\
        8 & disgusted \\
        9 & ashamed, guilty, embarrassed \\
        10 & devastated, sad, disappointed, nostalgic, lonely \\
        11 & surprised \\
        12 & impressed \\
        13 & sentimental \\
        14 & neutral \\
        15 & agreeing, acknowledging \\
        16 & encouraging \\
        17 & consoling, sympathizing \\
        18 & suggesting \\
        19 & questioning \\
        20 & wishing \\
        \bottomrule
    \end{tabular}
    \caption{Group of similar emotions and intents.}
    \label{tab:group_sim_emo_intent}
\end{table}

\end{document}